\documentclass[twoside]{article}

\usepackage[accepted]{arxiv}

\usepackage[round]{natbib}

\usepackage{xcolor}
\usepackage{times}  
\usepackage{helvet}  
\usepackage{courier}  
\usepackage[hyphens]{url}  
\usepackage{graphicx} 
\usepackage{caption} 

\usepackage{amsmath,bbm,esint}
\usepackage{amssymb}
\usepackage[toc,page]{appendix}
\usepackage[ruled,vlined]{algorithm2e}
\usepackage{caption}
\usepackage{subcaption}
\newtheorem{proposition}{Proposition}
\newtheorem{theorem}{Theorem}

\graphicspath{{./graphics/}}
\usepackage{algorithm2e}
\usepackage{enumitem}

\begin{document}

%

%

\makeatletter
\newcommand{\printfnsymbol}[1]{%
  \textsuperscript{\@fnsymbol{#1}}%
}
\makeatother

\twocolumn[


\aistatstitle{Training Wasserstein GANs without gradient penalties}

\aistatsauthor{ 
Dohyun Kwon \textsuperscript{\rm 1} \printfnsymbol{1},
    Yeoneung Kim \textsuperscript{\rm 2} \printfnsymbol{1},
    Guido Mont{\'u}far \textsuperscript{\rm 3},
    Insoon Yang \textsuperscript{\rm 2}
}

\aistatsaddress{ 
\textsuperscript{\rm 1} University of Wisconsin-Madison \And
    \textsuperscript{\rm 2} Seoul National University \And
    \textsuperscript{\rm 3} UCLA / Max Planck Institute MIS} ]

\begin{abstract}
 We propose a stable method to train Wasserstein generative adversarial networks. In order to enhance stability, we consider two objective functions using the $c$-transform based on Kantorovich duality which arises in the theory of optimal transport. We experimentally show that this algorithm can effectively enforce the Lipschitz constraint on the discriminator while other standard methods fail to do so. As a consequence, our method yields an accurate estimation for the optimal discriminator and also for the Wasserstein distance between the true distribution and the generated one. Our method requires no gradient penalties nor corresponding hyperparameter tuning and is computationally more efficient than other methods. At the same time, it yields competitive generators of synthetic images based on the MNIST, F-MNIST, and CIFAR-10 datasets. 
\end{abstract}

\section{Introduction}
	Generative Adversarial Networks (GANs) \citep{Goo14} have seen remarkable success in generating synthetic images. In GANs there are two networks, the generator and the discriminator, that compete with each other. Accordingly, the procedure seeks to find a minimax solution, Nash equilibrium. It has found numerous applications in machine learning, including semisupervised learning and speech synthesis \citep{Don19}. However, training GANs remains a difficult problem due to intrinsic instability associated with mode collapse and gradient vanishing \citep{GAAD17}. Several works have sought to improve the stability \citep{Tol17, Sal16, Now16, Rad15,kodali2017,liu2017} and have substantially improved beyond the original GAN. 
	Of particular interest in this context is the Wasserstein GAN (WGAN) framework 
	proposed by \citet{Arj17}, where the training objective for the generator network is the Wasserstein distance to the target distribution. 
	
	The possible advantages of WGANs over GANs in terms of performance and convergence have been demonstrated experimentally \citep[see][]{Arj17} but their mathematical properties are yet to be understood thoroughly. In WGANs one seeks to solve a minimax problem over the generator and the discriminator with the discriminator 
	restricted to be a 1-Lipschitz function. In the original WGAN, 
	weight clipping was introduced in order to approximately implement the 1-Lipschitz condition, but this 
	can 
	overly restrict the class of functions. To overcome this issue, WGANs with gradient penalty (WGAN-GP) were proposed by \citet{GAAD17}. This approach relies on the fact that the gradient norm of an optimal discriminator $\phi$ is 1 almost everywhere, i.e.\ $\|D\phi\|=1$. This requires explicit computation of the gradient of the discriminator at sample points and interpolated sample points. \citet{Wei18} pointed out that applying the gradient penalty only at sample points is insufficient. Further, \citet{Pet18} emphasized that $\|D\phi\|=1$ is not necessarily satisfied globally 
	and requiring it can harm the optimization. Therefore, they proposed to consider a manifold around sample points and penalize only gradient norms above 1 to approximate the Wasserstein distance in a stable manner. 
	However, their method still depends on forward and backward computation of neural networks to compute the gradients, which is computationally costly. 
	
	In this paper, we investigate avenues to improve the stability, accuracy, and computational cost of WGAN training. First proposed by \citet{MMG19}, $c$-transform methods were demonstrated to allow for an accurate estimation of the true Wasserstein metric. However, these methods are unstable and do not perform very well in the generative setting (see our theoretical investigation in Section~\ref{sec:mo} and \citealt[][Table 2 and Figure 3]{MMG19}). 
	
	We propose a novel stable training method based on the $c$-transform with multiple objective functions, inspired by the back-and-forth method for the optimal transportation problem on spatial grids proposed by \citet{JacLeg19}. Usually the objective function for the generator in WGANs is defined based on the Kantorovich duality formulation of the 1-Wasserstein distance, which is given by 
	\begin{align}
		\label{eqn:kan}
		W_1(\mu,\nu) = \sup_{\phi \in Lip_1} \left( \mathbb{E}_\mu[\phi] - \mathbb{E}_\nu[\phi] \right), 
	\end{align}
	where $Lip_1$ is a set of Lipschitz continuous functions, i.e.\ functions satisfying $|\phi(x)-\phi(y) \leq |x-y|$. 
	Instead of this, we consider two optimization problems given as 
    \begin{equation*}
        \sup_{\phi \in C} \left( \mathbb{E}_\mu[\phi] + \mathbb{E}_\nu[\phi^c] \right)
        \text{ and }\sup_{\phi \in C} \left( \mathbb{E}_\mu[(-\phi)^c] + \mathbb{E}_\nu[-\phi] \right), 
    \end{equation*}
    where $C$ is simply a set of continuous functions and $\phi^c$ is the $c$-transform of $\phi$ that we will discuss in Section~\ref{sec:OToverview}. 
    %
    As we will see, our proposed method can solve both problems simultaneously. In addition, our algorithm also optimizes the usual Kantorovich duality formulation \eqref{eqn:kan}. 
	
    As noted above, 
    the WGAN with weight clipping has several complications, while WGAN-GP and other improved versions of WGANs require explicit computation of the discriminator gradients, 
    which represents computational costs and hyperparameter tuning. 
    In contrast, our method implements the constraint without the need to explicitly penalize the gradients. 
    Our contribution is summarized as: 
	\begin{itemize}[leftmargin=*]
		\item
		We propose a fast and stable WGAN training algorithm that enforces a 1-Lipschitz bound without the need of introducing a gradient penalty. Consequently, no hyperparameter tuning for such a penalty is needed.
		
		\item Our algorithm generates high-quality synthetic images and works well with various types of data, such as mixtures of Gaussians, MNIST, F-MNIST, CIFAR-10. 
		
		\item Our method not only computes the discriminator $\phi$ during 
		training but also accurately computes the 1-Wasserstein distance between target and generator. 
	\end{itemize}

\section{Theoretical Analysis}
	\subsection{Optimal transport overview}
	\label{sec:OToverview}
	
	For a bounded domain $\Omega \subset \mathbb{R}^d$, let $\mathcal{P}(\Omega)$ be the space of Borel probability measures on $\Omega$. The $p$-Wasserstein distance between two probability measures $\mu$ and $\nu$ in $\mathcal{P}(\Omega)$ is defined as
	\begin{align}
		\label{eqn:w}
		W_p(\mu,\nu) := \min \left\{ \int_{\Omega \times \Omega} |x-y|^p \text{d} \gamma : \gamma \in \Pi (\mu,\nu) \right\}.
	\end{align}
	Here, $\Pi (\mu,\nu)$ is the set of all Borel probability measures $\pi$ on $\Omega \times \Omega$ such that $\pi(A \times \Omega) = \mu(A) \hbox{ and } \pi(\Omega \times A) = \nu(A)$ for all measurable subsets $A \subset \Omega$. 
	The Kantorovich duality yields that $W_p(\mu,\nu)$ equals 
	\begin{align}
		\label{eqn:dual}
		\sup_{\phi, \psi \in C(\Omega)} \left\{ \int_\Omega \phi \text{d}\mu + \int_\Omega \psi \text{d}\nu : \phi(x) + \psi(y) \leq |x-y|^p \right\},
	\end{align}
	where $C(\Omega)$ is the set of bounded continuous functions on $\Omega$ (see, e.g.\ \citealt[Theorem 1.39]{San15} or \citealt[Theorem 5.9]{Vil08}). 
	It is well known that there exists a solution $(\phi, \psi)$ to \eqref{eqn:dual} and it has the form $\psi = \phi^c$, where $\phi^c$ is the $c$-transform of $\phi$, which is defined as $\phi^c(y) := \inf_{x \in \Omega} \left\{ |x-y|^p - \phi(x) \right\}$ for $y \in \Omega$ \citep[see][Proposition 1.11]{San15}. Notice that the above maximization problem \eqref{eqn:dual} is equivalent to 
	 \begin{align}
		\label{eqn:dual1}
		\sup_{\phi \in C(\Omega)} \left\{ \int_\Omega \phi \text{d}\mu + \int_\Omega \phi^c \text{d}\nu \right\}.
	\end{align}
	
	 For $p=1$, $\psi \in Lip_1$ if and only if there exists $\phi$ such that $\psi = \phi^c$. In particular, for all $\psi \in Lip_1$, we have $\phi^c = - \phi$. This leads to another dual formula of $W_1$, namely 
	 \begin{align}
		\label{eqn:dual2}
		W_1(\mu,\nu) = \sup_{\phi \in Lip_1} \left\{ \int_\Omega \phi \text{d}(\mu - \nu)\right\},
	\end{align}
	which is widely used particularly in WGANs. Here $\phi$ plays the role of discriminator and $\nu$ is the generative distribution. The objective of WGAN is to find $\nu$ from a parametrized set of pushforward maps of a noise source that best mimics the $\mu$ which is given. In practice, the integral is replaced by a sample average and optimization is conducted iteratively using mini-batches of samples.
	
    \subsection{The universal admissible condition}

    To find a probability measure $\nu$ that best mimics $\mu$ using mini-batches, we first define the so-called universal admissible set as the set of $\phi$'s which satisfy
    \begin{align}
    \label{eqn:con}
    \phi(x) - \phi(y) \leq |x-y| \hbox{ for all } (x,y) \in \text{supp}(\mu) \times \text{supp}(\nu).
    \end{align}
    If both $\text{supp}(\mu)$ and $\text{supp}(\nu)$ are equal to $\Omega$, then \eqref{eqn:con} is equivalent to the 1-Lipschitzness on $\Omega$, which rarely happens in real-world data.
    
    On the other hand, any 1-Lipschitz function on $\Omega$ satisfies the above condition. Thus, our condition is weaker than the 1-Lipschitzness on $\Omega$. Using the duality formulation, we prove that this relaxation still computes the Wasserstein distance.

    \begin{theorem}
    \label{thm:weak}
    For any probability measures $\mu$ and $\nu$, we have
    \begin{align}
    		\label{eqn:weak}
    		W_1(\mu,\nu) = 
    		\sup\left\{ \int_\Omega \phi \text{d}(\mu - \nu) : \phi \hbox{ satisfies } \eqref{eqn:con} \right\}.
    	\end{align}
    \end{theorem}
    
    \textit{Proof.} Recall the dual formulation \eqref{eqn:dual2} of $W_1$. As all Lipschitz functions satisfy \eqref{eqn:con}, we have
    \begin{align*}
        W_1(\mu, \nu) \leq \sup\left\{ \int_\Omega \phi \text{d}(\mu - \nu) : \phi \hbox{ satisfies } \eqref{eqn:con} \right\}.
    \end{align*}
    
    On the other hand, consider the so-called transport plan $\gamma : \Omega \times \Omega \rightarrow \mathbb{R}$, which satisfies 
    \begin{align}
    \label{eqn:tpp}
    \gamma(A \times \Omega) = \mu(A) \hbox{ and } \gamma(\Omega \times A) = \nu(A)    
    \end{align}
    for all measurable subsets $A \subset \Omega$. As $\text{supp}(\gamma) \subset \text{supp}(\mu) \times \text{supp}(\nu)$, we have 
    \begin{align*}
    \int_\Omega \phi (\text{d}\mu - \text{d}\nu) &= \int_{\Omega \times \Omega} \phi(x) - \phi(y) \text{d} \gamma(x,y),\\ 
    &\leq \int_{\Omega \times \Omega} |x-y| \text{d} \gamma 
    \end{align*}
    for all $\phi$ satisfying \eqref{eqn:con} and any transport plan $\gamma$ satisfying \eqref{eqn:tpp}. 
    As a consequence of 
    this inequality and \eqref{eqn:w}, we have
    \begin{align*}
    &\sup\left\{ \int_\Omega \phi (\text{d}\mu - \text{d}\nu) : \phi \hbox{ satisfies } \eqref{eqn:con} \right\}\\ &\leq \inf_{\gamma \in \Pi (\mu,\nu)} \left\{ \int_{\Omega \times \Omega} |x-y| \text{d} \gamma(x,y) \right\} = W_1(\mu, \nu).
    \end{align*} 
    Thus, we conclude \eqref{eqn:weak}. \hfill$\Box$
    
    Consequently, instead of the Lipschitz condition on $\Omega$, it suffices to enforce \eqref{eqn:con} while training WGANs. This observation plays an important role in our Algorithm 1. 

\subsection{Comparison between objective functions}

    In our approach 
    we optimize multiple objective functions simultaneously. They originate from two duality formulas, \eqref{eqn:dual1} and \eqref{eqn:dual2},
	\begin{align*}
		\mathcal{J}_1(\phi) &= \int_\Omega \phi \text{d}\mu_n + \int_\Omega (-\phi) \text{d}\nu_n, \\
		\mathcal{J}_2(\phi) &= \int_\Omega \phi \text{d}\mu_n + \int_\Omega \phi^c(\cdot ; \mu_n) \text{d}\nu_n,\\
		\mathcal{J}_3(\phi) &= \int_\Omega (-\phi)^c (\cdot ; \nu_n) \text{d}\mu_n + \int_\Omega (-\phi) \text{d}\nu_n,\\
		\mathcal{J}_4(\phi) &= \int_\Omega (-\phi)^c (\cdot ; \nu_n) \text{d}\mu_n + \int_\Omega \phi^c (\cdot ; \mu_n) \text{d}\nu_n.
	\end{align*}
	
	Here, we consider empirical distributions $\mu_n$ and $\nu_n$ sampled from probability measures $\mu$ and $\nu$, respectively, as in \eqref{eqn:emp}. Then, we shall take the infimum in the $c$-transform over the support of the probability measure instead of 
	$\Omega$. For $\phi : \Omega \to \mathbb{R}$ and $\eta \in \mathcal{P}(\Omega)$, the $c$-transform on the support of $\eta$ is a function $\phi^c(\cdot; \eta) : \Omega \to \mathbb{R}$ given by 
	\begin{align*}
		\phi^c(y ; \eta) := \inf_{x \in \text{supp}(\eta)} \left\{ c(x,y) - \phi(x) \right\} \hbox{ for } y \in \Omega.
	\end{align*}
	This concept has been implicitly used in the literature, e.g.\ in the works of \cite{farnia2020gans, MMG19}. However, to our knowledge, the discrepancy between $\phi^c$ and $\phi^c(\cdot ; \eta)$ 
	has not been studied in detail. 

	The original $c$-transform satisfies $\phi^c \leq -\phi$ over $\Omega$, but this relation does not hold for $\phi^c ( \cdot ; \eta)$ in general. In turn, $\phi^c$ is not necessarily equal to $-\phi$ even if $\phi$ is a 1-Lipschitz function. However, the following inequalities hold for the $\mathcal{J}_i$'s. 
	
	\begin{theorem}
	\label{thm:cp1}
		If $\phi$ satisfies the admissibility property \eqref{eqn:con}, then we have
		\begin{align}
		\label{eqn:js}
			\mathcal{J}_1 (\phi) \leq \mathcal{J}_2 (\phi)  \leq \mathcal{J}_4(\phi) \hbox{ and } \mathcal{J}_1 (\phi) \leq \mathcal{J}_3 (\phi)  \leq \mathcal{J}_4(\phi).
		\end{align}
	\end{theorem}
    \textit{Proof.}
	From \eqref{eqn:con} and $\text{supp}(\mu_n) \subset \text{supp}(\mu)$, it holds that for all $y \in \Omega$
	\begin{align*}
		\phi^c(y ; \mu_n) &:= \inf_{x \in \text{supp}(\mu_n)} \left\{ |x-y| - \phi(x) \right\},\\ &\geq \inf_{x \in \text{supp}(\mu_n)} \left\{ - \phi(y) \right\} = - \phi(y).
	\end{align*}
	Therefore, we conclude $\mathcal{J}_1 (\phi) \leq \mathcal{J}_2 (\phi)$. The other inequalities can be shown similarly. 
	\hfill$\Box$
	
    The above inequalities give us an important observation or test on the enforcement of \eqref{eqn:con}. In particular, equivalence between the inequalities and \eqref{eqn:con} holds as follows. The theorem below still holds if we replace $\mathcal{J}_1(\phi) \leq \mathcal{J}_2(\phi)$ by other suitable inequalities, for instance, $\mathcal{J}_1(\phi) \leq \mathcal{J}_3(\phi)$.
    
    \begin{theorem}
    \label{thm:cp2}
    If $\mathcal{J}_1(\phi) \leq \mathcal{J}_2(\phi)$ for all $\mu_n$ and $\nu_n$, then $\phi$ satisfies the admissibility property \eqref{eqn:con}. Here, $\mu_n$ and $\nu_n$ are empirical measures from $\mu$ and $\nu$ 
    as given in \eqref{eqn:emp}. 
    \end{theorem}
    
    \textit{Proof.} For any $x \in \text{supp}(\mu)$ and $y \in \text{supp}(\nu)$, if $X_i = x$  and $Y_i = y$ for all $i=1,2,\dots, n$, then the empirical measures are $\mu_n = \delta_x$ and $\nu_n = \delta_y$. 
	From $J_1(\phi) \leq J_2(\phi)$, we have $$\phi(x) - \phi(y) \leq \phi(x) + \phi(y; \delta_x).$$
	As $\phi(y; \delta_x) = |x-y| - \phi(x)$, we conclude
	$$\phi(x) - \phi(y) \leq |x-y|.$$
	Hence \eqref{eqn:con} holds for any $x \in \text{supp}(\mu)$ and 
	$y \in \text{supp}(\nu)$. \hfill$\Box$

    Motivated by the above theorems, in the next section we construct an algorithm that reduces the distance between the $\mathcal{J}_i$'s. This way, we can build a 1-Lipschitz continuous function $\phi$ that better approximates the Wasserstein distance between the true distributions using mini-batches. 

\section{Our Algorithm}

\subsection{Main algorithm}

    Now we go back to \eqref{eqn:weak} and minimize it with respect to $\nu \in \mathcal{P}(\Omega)$: 
    \begin{align}
		\label{eqn:dual3}
		\inf_{\nu \in P(\Omega)}
		\sup\left\{ \int_\Omega \phi \text{d}(\mu - \nu) : \phi \hbox{ satisfies } \eqref{eqn:con} \right\}.
	\end{align}
    Based on the observation in Theorems~\ref{thm:cp1} and \ref{thm:cp2}, we propose the following Algorithm~\ref{alg:1}. In the 
    algorithm, $\mathcal{J}_2$ or $\mathcal{J}_3$ is updated only if the inequality \eqref{eqn:js} is not satisfied. Furthermore, as the objective function in \eqref{eqn:dual3} is $\mathcal{J}_1$, we do not need to optimize $\mathcal{J}_2$ nor $\mathcal{J}_3$ when training the generator. Since the algorithm depends on the comparison between $\mathcal{J}_i$'s, we call it \textbf{CoWGAN}. We parametrize the discriminator $\phi$ by $\eta$ and the generator network by $\theta$. 
    We used Adam in our experiments, but this can be replaced by any other optimizer. 
	
	\begin{algorithm}
	\label{al0}
	\For{$iter$ of training iterations}{
		\For{$t = 1, 2, \dots, N_{critic}$}{
			\uIf{$\mathcal{J}_2 < \mathcal{J}_1$ }{
				$\eta \leftarrow \hbox{Adam}(-\mathcal{J}_2, \eta)$
			}
			\uElseIf{$\mathcal{J}_3 < \mathcal{J}_1$}{
				$\eta \leftarrow \hbox{Adam}(-\mathcal{J}_3, \eta)$
			}
			\Else{
				$\eta \leftarrow \hbox{Adam}(-\mathcal{J}_1, \eta)$
				}
		}
	$\theta \leftarrow  \hbox{Adam}(\mathcal{J}_1, \theta)$  
	}	
		\caption{Our proposed algorithm (\textbf{CoWGAN}) to find the minimum of  \eqref{eqn:dual3}}
		\label{alg:1}
	\end{algorithm}

\subsection{Why does our algorithm work?} 
The above algorithm can be understood as a simple indirect way to have \eqref{eqn:con} satisfied. From Theorem 1, if $\mathcal{J}_2 < \mathcal{J}_1$ or $\mathcal{J}_3 < \mathcal{J}_1$, then there exists a pair $(x,y) \in \text{supp}(\mu) \times \text{supp}(\nu)$ such that \eqref{eqn:con} does not hold.
    
    Then, formally, the condition \eqref{eqn:con} will be enforced through the following procedure. Assume that $\phi(x) - \phi(y) - |x-y|$ is much larger than zero for some $x \in \text{supp}(\mu)$  and $y \in \text{supp}(\nu)$ and thus $(x,y)$ does not satisfy \eqref{eqn:con}. When $\mathcal{J}_2$ is optimized, $\phi(x)$ decreases due to $\phi^c(y ; \mu) = |x-y| - \phi(x)$. On the other hand, when $\mathcal{J}_3$ is optimized, $\phi(y)$ increases due to $(-\phi)^c(x ; \nu) = |x-y| - \phi(y)$. As a consequence, $\phi(x) - \phi(y) - |x-y|$ decreases after the iteration. 
    
    In Section~\ref{sec:estimatingW} we will experimentally show that even with a very small mini-batch size $n$, our algorithm can enforce the condition~\eqref{eqn:con} effectively. 
    In the next two subsections 
    we 
    discuss why we do not use $\mathcal{J}_2$ or $\mathcal{J}_3$ as objective functions but rather $\mathcal{J}_1$, and 
    why other methods based on $\mathcal{J}_2$ may lead to poor generators. 

\subsection{Revisit of the Kantorovich duality}
\label{sec:3.3}
	
    In practice, one does not have access to the true distribution, but rather to mini-batches that are sampled from the available training data set. We observe that the objective functions $\mathcal{J}_2$ and $\mathcal{J}_3$ may not be suitable for the construction of the true distribution. This is why $\mathcal{J}_2$ and $\mathcal{J}_3$ are excluded and $\mathcal{J}_1$ is only used when updating the generator network in Algorithm 1.
    
    To illustrate this, we first consider $n$ i.i.d.\ observations $X_1, X_2, \ldots, X_n$ distributed according to $\mu$, and $n$ i.i.d.\ observations $Y_1, Y_2,\ldots,Y_n$ distributed according to $\nu$. Let $\mu_n$ and $\nu_n$ be the empirical measures based on $X_i$'s and $Y_i$'s, respectively, 
    \begin{align}
    \label{eqn:emp}
    \mu_n := \frac{1}{n} \sum_{i=1}^n \delta_{X_i} \hbox{ and }
    \nu_n := \frac{1}{n} \sum_{i=1}^n \delta_{Y_i}.
    \end{align}
    Let $\mathcal{P}_n(\Omega)$ denote a set of empirical measures supported on at most $n$ points in $\Omega$. 
    
    Recall that training WGANs aims to minimize \eqref{eqn:dual2} using mini-batches, which formally means to solve 
    \begin{align}
    \label{min1}
    \inf_{\nu \in P(\Omega)} \sup_{\phi \in Lip_1} \mathbb{E}_{\mu_n \sim \mu, \nu_n \sim \nu} \left[  \int_\Omega \phi \text{d}(\mu_n - \nu_n)\right].
    \end{align} 
    Here, $\mu_n$ is a mini-batch from the dataset and $\nu_n$ is a discrete measure from the generator. As $\int_\Omega \phi \text{d}(\mu_n - \nu_n)$ is linear with respect to $\mu_n$ and $\nu_n$, one can easily check that the above formula is equivalent to the minimization problem of \eqref{eqn:dual2} with respect to $\nu$.
    
    However, the $c$-transforms on mini-batches, $\phi^c(\cdot ; \mu_n)$ and $(-\phi)^c(\cdot ; \nu_n)$ in $\mathcal{J}_2$ and $\mathcal{J}_3$, depend on $\mu_n$ and $
    \nu_n$ respectively. Therefore, $\mathcal{J}_2$ and $\mathcal{J}_3$ are not linear with respect to $\mu$ or $\nu$. As a consequence, the minimization problem of \eqref{eqn:dual2} with respect to $\nu$ cannot be solved by 
    considering 
    \begin{align}
    \label{min2}
    \inf_{\nu \in P(\Omega)} \sup_{\phi} \mathbb{E}_{\mu_n \sim \mu, \nu_n \sim \nu} \left[  \mathcal{J}_2 (\text{ or } \mathcal{J}_3) \right].
    \end{align} 

\subsection{Mini-batch optimal transport}
\label{sec:mo}

    Interchanging expectation and supremum in \eqref{min2}, we get the following problem:
    \begin{align}
    \label{min3}
    \inf_{\nu \in P(\Omega)} \mathbb{E}_{\mu_n \sim \mu, \nu_n \sim \nu} [ W_1(\mu_n, \nu_n)].
    \end{align} 
    The question is if an optimal $\nu$ in \eqref{min3} is similar to the given probability measure $\mu$. The answer is \emph{no} as illustrated in the following proposition.
    \begin{proposition}
    \label{prop:1}
    Assume that $d=n=1$ and $\mu \in \mathcal{P}_m(\Omega)$ for $m>1$. Then, for any median $y$ of $\mu$, $\nu = \delta_{y}$ is a global minimizer of \eqref{min3}.
    Here, a median of a probability measure $\mu$ is defined as a real number $k$ satisfying $\mu((-\infty, k]) \geq \frac{1}{2}$ and $\mu([k, \infty)) \geq \frac{1}{2}$.
    \end{proposition}
    
    The proof of Proposition~\ref{prop:1} can be found in the supplementary material. The result 
    implies that $\nu$ might completely fail to mimic $\mu$ using mini-batches when $\mu$ is a discrete measure. This 
    observation coincides with the blurry images obtained in previous works using the $c$-transform method to train WGANs \citep{MMG19}. 
    
    In general, the discrepancy between the true distance and the empirical one has been studied by \cite{Arora17, Arora18, Bai18}.  Furthermore, in higher dimensions, even if $\nu_n$ comes from $\mu$ and even if the latter is a normal distribution, $W_1(\mu_n,\nu_n)\geq C>0$ for some $C$ where $\mu_n$ and $\nu_n$ are discrete probability measures from $\mu$; see the works of \citet{Fat20,Fat21} for details.

\section{Estimating the 1-Wasserstein distance} 
\label{sec:estimatingW}

In this section we give an empirical demonstration of how our algorithm can compute the 1-Wasserstein distance. 

\subsection{Synthetic datasets in 2D}

    On synthetic datasets in 2D, we observe that the condition \eqref{eqn:con} can be efficiently enforced based on our comparison method (CoWGAN). See Algorithm~\ref{al1}. Furthermore, this method can achieve 
    \begin{align}
    \label{eqq}
    \mathcal{J}_1 = \mathcal{J}_2 = \mathcal{J}_3 = \mathcal{J}_4    
    \end{align}
    if the size of mini-batches is sufficiently large.

    \begin{algorithm}[t]
	
		\For{$t = 1, 2, \dots, N_{critic}$}{
			\uIf{$\mathcal{J}_2 < \mathcal{J}_1$}{
				$\eta \leftarrow \hbox{Adam}(-\mathcal{J}_2, \eta)$
			}
			\uElseIf{$\mathcal{J}_3 < \mathcal{J}_1$}{
				$\eta \leftarrow \hbox{Adam}(-\mathcal{J}_3, \eta)$
			}
			\Else{
				$\eta \leftarrow \hbox{Adam}(-\mathcal{J}_1, \eta)$
				}
		}
		\caption{Our proposed algorithm to optimize \eqref{eqn:weak}}
	\label{al1}
	\end{algorithm}

    \begin{figure}[t]
    \centering
    \begin{tabular}{cc}
    CoWGAN (ours) & $c$-transform\\ 
    \includegraphics[width=0.23\textwidth,clip=true,trim=0cm 0cm 0cm 1cm  ]{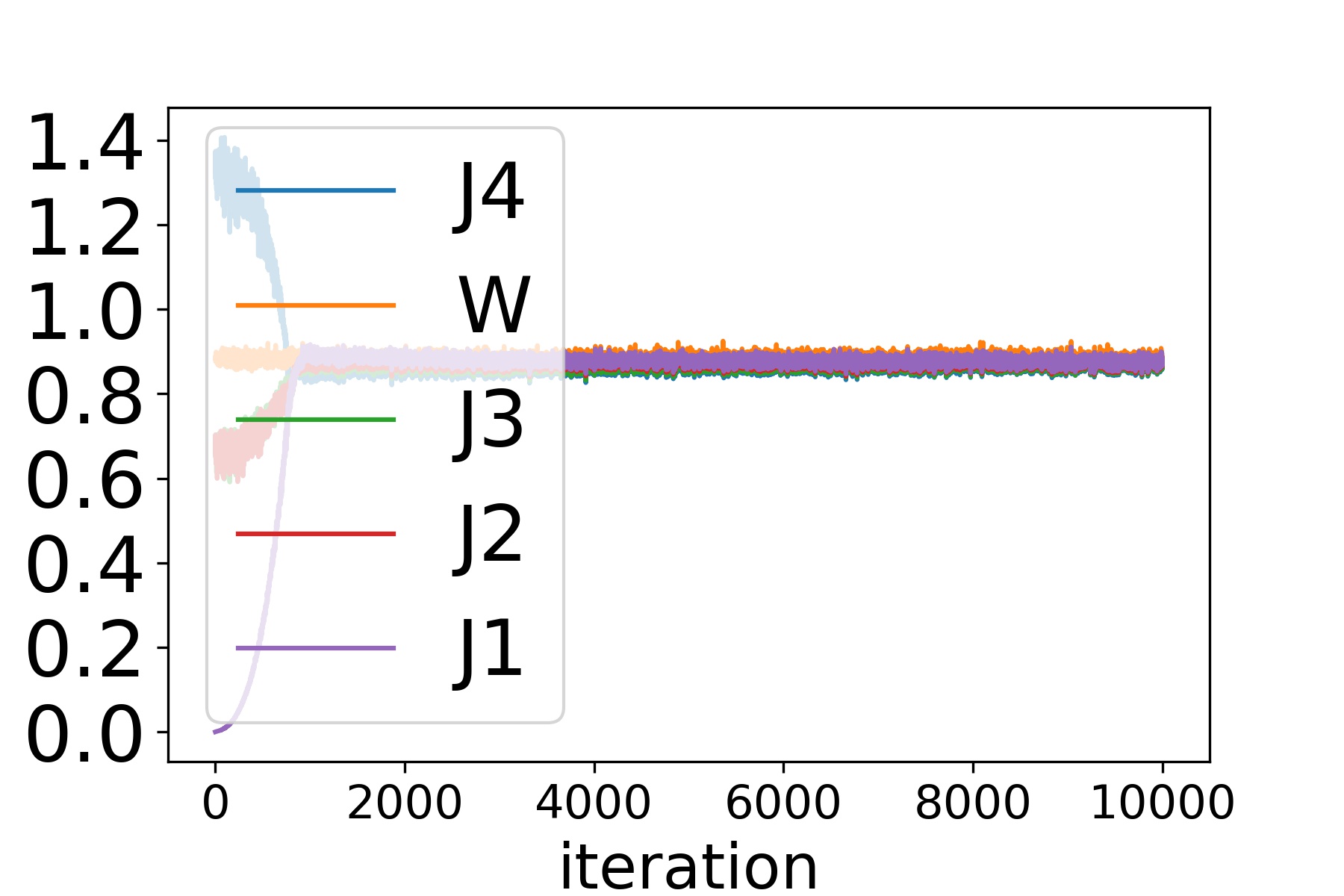}&
    \includegraphics[width=0.23\textwidth,clip=true,trim=0cm 0cm 0cm 1cm  ]{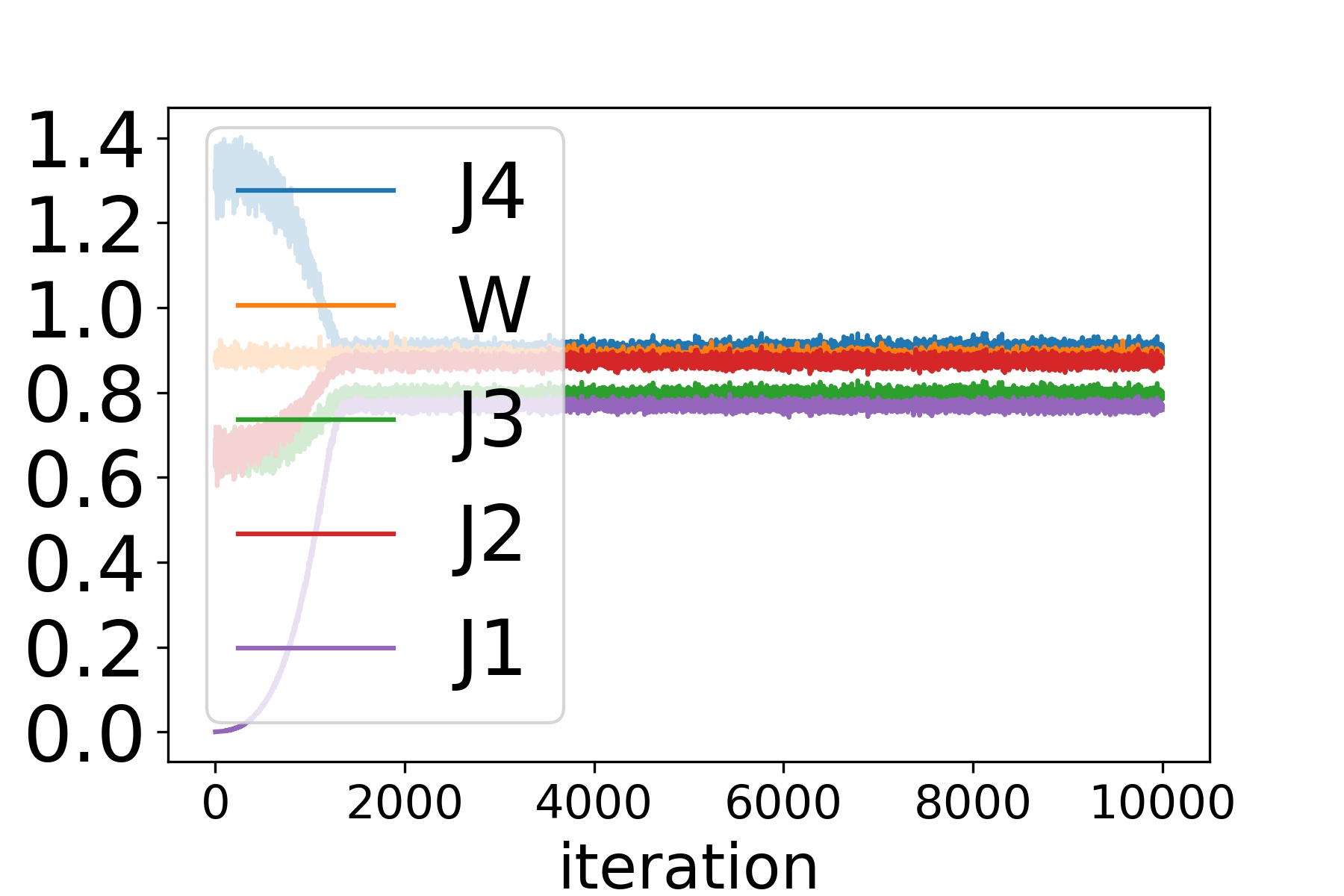}
    \\
    WGAN-GP, $\lambda=1$ & WGAN-GP, $\lambda=10$\\
    \includegraphics[width=0.23\textwidth,clip=true,trim=0cm 0cm 0cm 1cm  ]{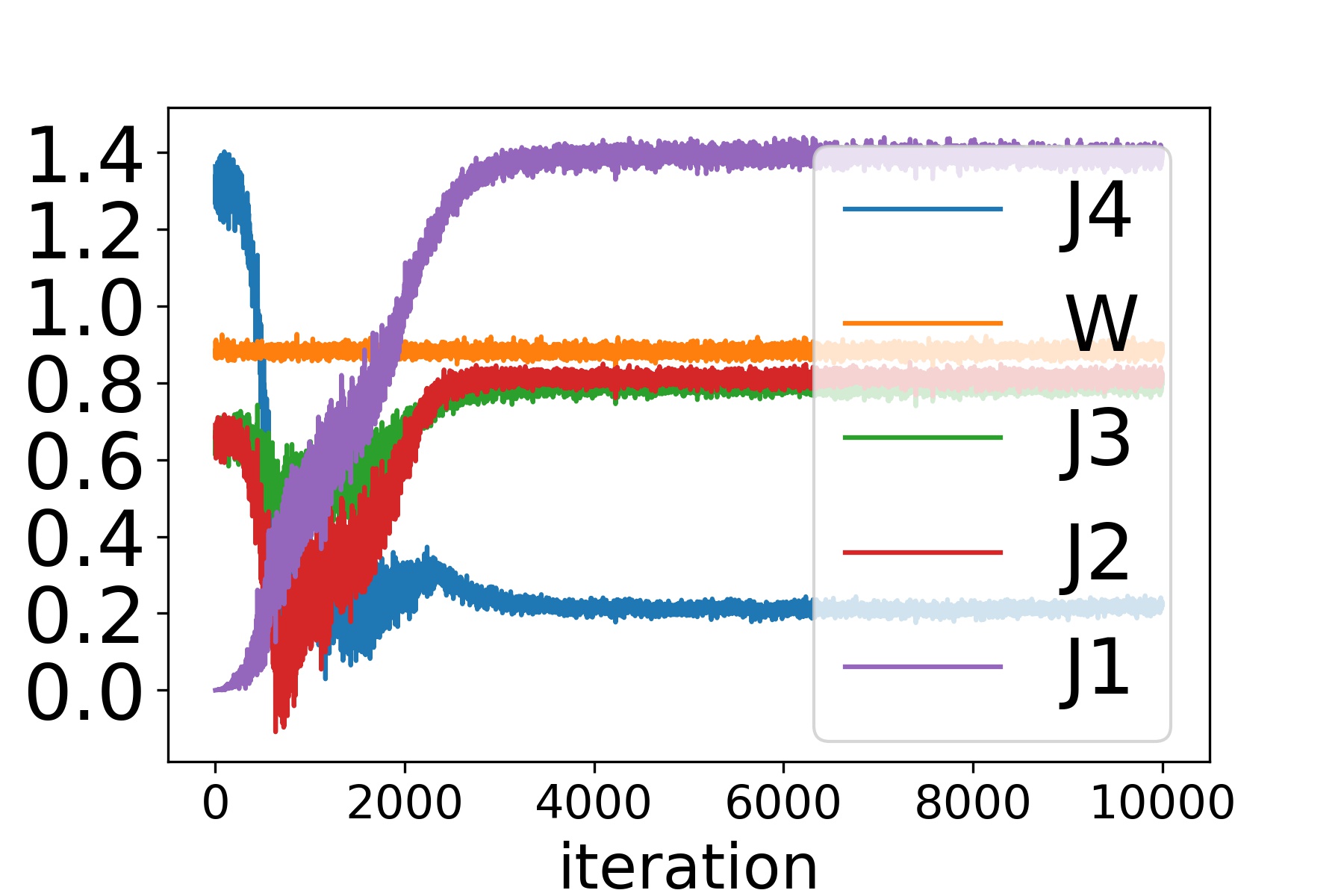}&
    \includegraphics[width=0.23\textwidth,clip=true,trim=0cm 0cm 0cm 1cm  ]{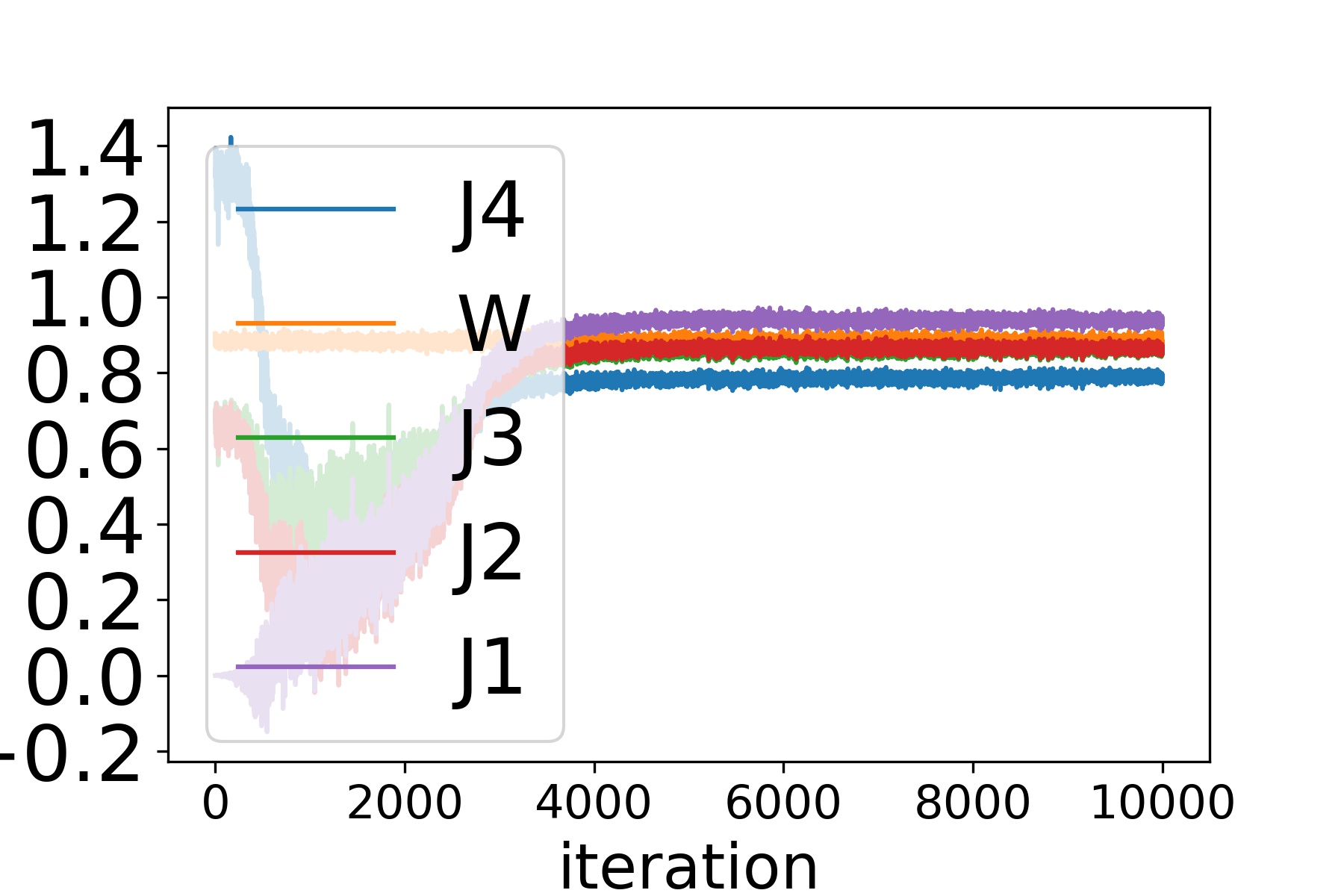}
    \end{tabular}
    \caption{The $\mathcal{J}_i$'s and the true Wasserstein distance (W).}
    \label{ot_comparison}
    \end{figure}
	
\begin{figure}[t]
    \centering
    \begin{tabular}{cc}
    CoWGAN (ours) & $c$-transform\\     
        \includegraphics[width=0.23\textwidth,clip=true,trim=0cm 0cm 0cm 1cm   ]{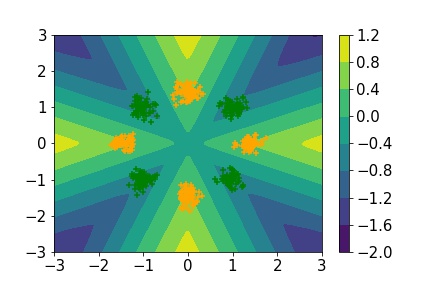}&
        \includegraphics[width=0.23\textwidth,clip=true,trim=0cm 0cm 0cm 1cm   ]{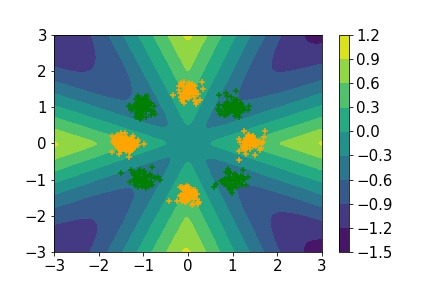}
        \\
    WGAN-GP, $\lambda=1$ & WGAN-GP, $\lambda=10$\\
        \includegraphics[width=0.23\textwidth,clip=true,trim=0cm 0cm 0cm 1cm   ]{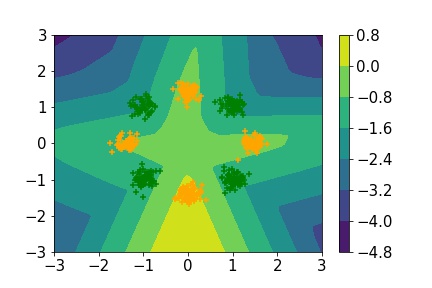}&
        \includegraphics[width=0.23\textwidth,clip=true,trim=0cm 0cm 0cm 1cm   ]{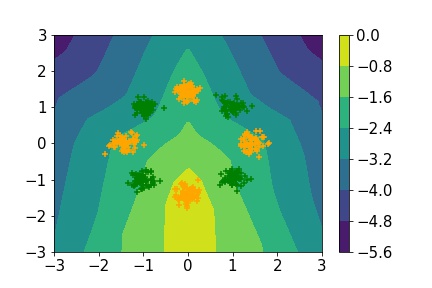}
\end{tabular}        
    \caption{The discriminator $\phi$ for two mixtures of 4 Gaussians (samples shown as green and yellow dots) after 2000 iterations with different methods and mini-batch size 256. 
    }    
    \label{fig:ex1}
    \end{figure}

    In Figure~\ref{ot_comparison}, we compute the Wasserstein distance between two different distributions. Each distribution is 
    a mixture of 4 different Gaussian distributions. See yellow points and green points in Figure~\ref{fig:ex1}. As shown in Figure~\ref{ot_comparison}, CoWGAN converges much faster than the other algorithms tested, namely the $c$-transform method \citep{MMG19}, and WGAN-GP with $\lambda = 1$ or $\lambda = 10$ \citep{GAAD17}. In particular, \eqref{eqq} does not hold in these other methods. 
	
\begin{figure}[t]
    \centering
        \begin{tabular}{cc}
    CoWGAN (ours) & $c$-transform\\     
        \includegraphics[width=0.23\textwidth,clip=true,trim=0cm 0cm 0cm 1cm    ]{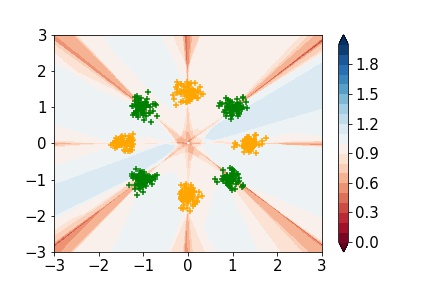}&
        \includegraphics[width=0.23\textwidth,clip=true,trim=0cm 0cm 0cm 1cm    ]{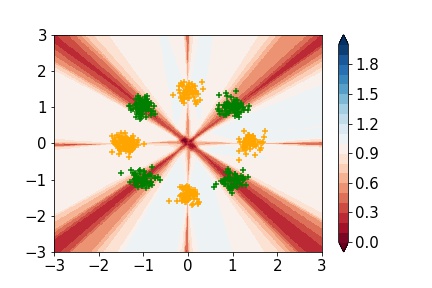}
        \\
            WGAN-GP, $\lambda=1$ & WGAN-GP, $\lambda=10$\\
        \includegraphics[width=0.23\textwidth,clip=true,trim=0cm 0cm 0cm 1cm    ]{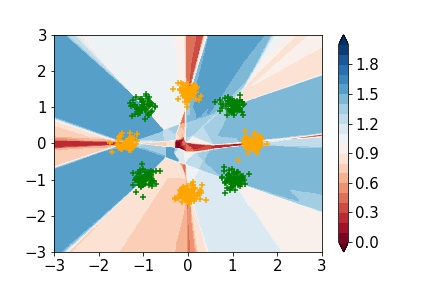}&
        \includegraphics[width=0.23\textwidth,clip=true,trim=0cm 0cm 0cm 1cm    ]{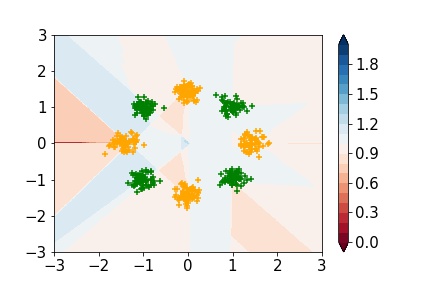}
\end{tabular}                
    \caption{Complementary to Figure~\ref{fig:ex1}. Shown is the gradient norm of the discriminator $\|D\phi\|$ for two mixtures of 4 Gaussians after 2000 iterations with different methods and mini-batch size 256. 
    }
    \label{fig:ex2}
    \end{figure}

    The discriminator $\phi$ obtained by Algorithm~\ref{al1} is more accurate with the same training time as shown in Figures~\ref{fig:ex1} and \ref{fig:ex2}. In particular, we observe that in our algorithm $\|D\phi\| = 1$ a.e.\ and the Lipschitzness of $\phi$ is enforced well. 
    The optimal discriminator should have level sets orthogonal to the transport map, which is approximately true in the top row but not in the bottom row of Figure~\ref{fig:ex1}.

\subsection{
Lipschitz constraint with 
small mini-batches} 

\begin{figure}[t]
    \centering
    \begin{tabular}{cc}
    CoWGAN (ours) & $c$-transform\\     
    \includegraphics[width=0.23\textwidth,clip=true,trim=0cm 0cm 0cm 1cm ]{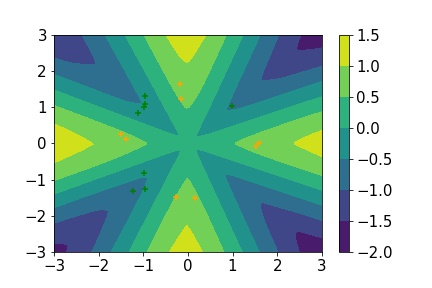}&
    \includegraphics[width=0.23\textwidth,clip=true,trim=0cm 0cm 0cm 1cm ]{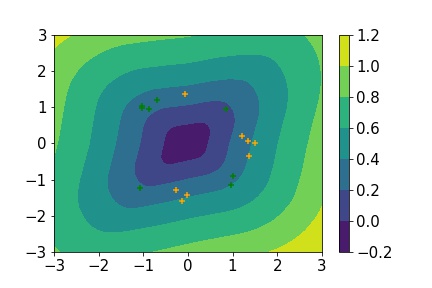}
    \\
    WGAN-GP, $\lambda=1$ & WGAN-GP, $\lambda=10$\\    
    \includegraphics[width=0.23\textwidth,clip=true,trim=0cm 0cm 0cm 1cm ]{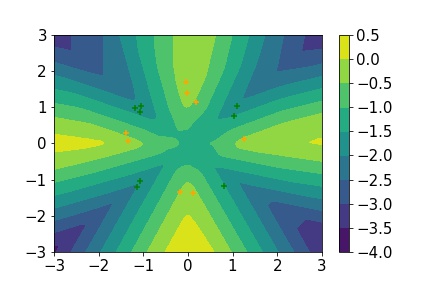}&
    \includegraphics[width=0.23\textwidth,clip=true,trim=0cm 0cm 0cm 1cm ]{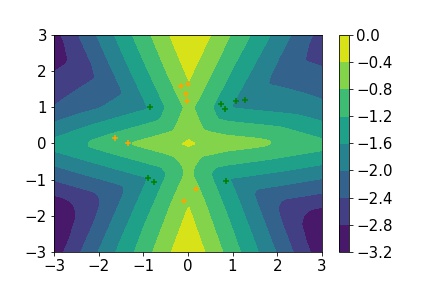}
    \end{tabular}
    \caption{The discriminator $\phi$ after 10,000 iterations with mini-batches of size 8. 
    }
    \label{fig:small1}
    \end{figure}

\begin{figure}[t]
    \centering
    \begin{tabular}{cc}
    CoWGAN (ours) & $c$-transform\\         
    \includegraphics[width=0.23\textwidth,clip=true,trim=0cm 0cm 0cm 1cm  ]{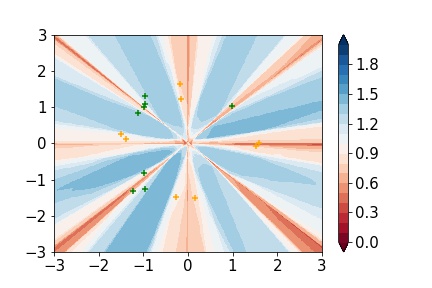} &
    \includegraphics[width=0.23\textwidth,clip=true,trim=0cm 0cm 0cm 1cm  ]{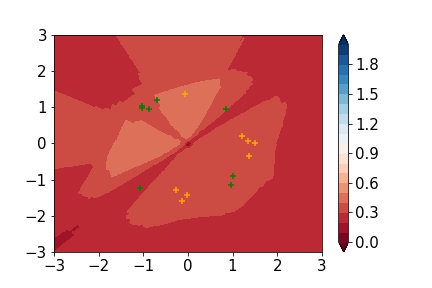}
    \\
    WGAN-GP, $\lambda=1$ & WGAN-GP, $\lambda=10$\\        
    \includegraphics[width=0.23\textwidth,clip=true,trim=0cm 0cm 0cm 1cm  ]{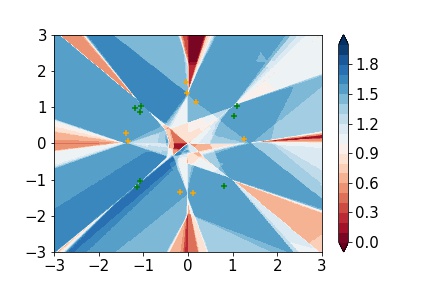}&
    \includegraphics[width=0.23\textwidth,clip=true,trim=0cm 0cm 0cm 1cm  ]{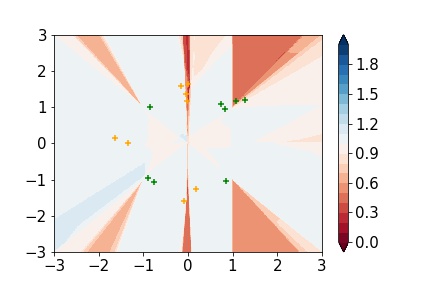}
    \end{tabular}
    \caption{Complementary to Figure~\ref{fig:small1}. Shown is $\|D\phi\|$ after 10,000 iterations with mini-batches of size 8. 
    }
    \label{fig:small2}
    \end{figure}

    Surprisingly, Algorithm~\ref{al1} can enforce the Lipschitz constraint very effectively even with 
    very small mini-batch size. We consider the same distributions 
    as in Figures~\ref{fig:ex1} and \ref{fig:ex2}, but reduce the size of the mini-batches to $8$. 
    Figures~\ref{fig:small1} and \ref{fig:small2} show that our algorithm still can accurately compute the optimal discriminator,  
whereas the other methods have significant 
difficulties 
even after more iterations. 

\subsection{The MNIST dataset}

    Computing the Wasserstein distance for high-dimensional data is not an easy task. We experimentally observe that CoWGAN can estimate the Wasserstein distance between 
    measures on a high-dimensional space efficiently. 
    
\begin{figure}[t]
    \label{fig:mnist_ot}
    \centering
    \begin{tabular}{cc}
    CoWGAN (ours) & $c$-transform \\  
    \includegraphics[width=0.23\textwidth,clip=true,trim=0cm 0cm 0cm 1cm ]{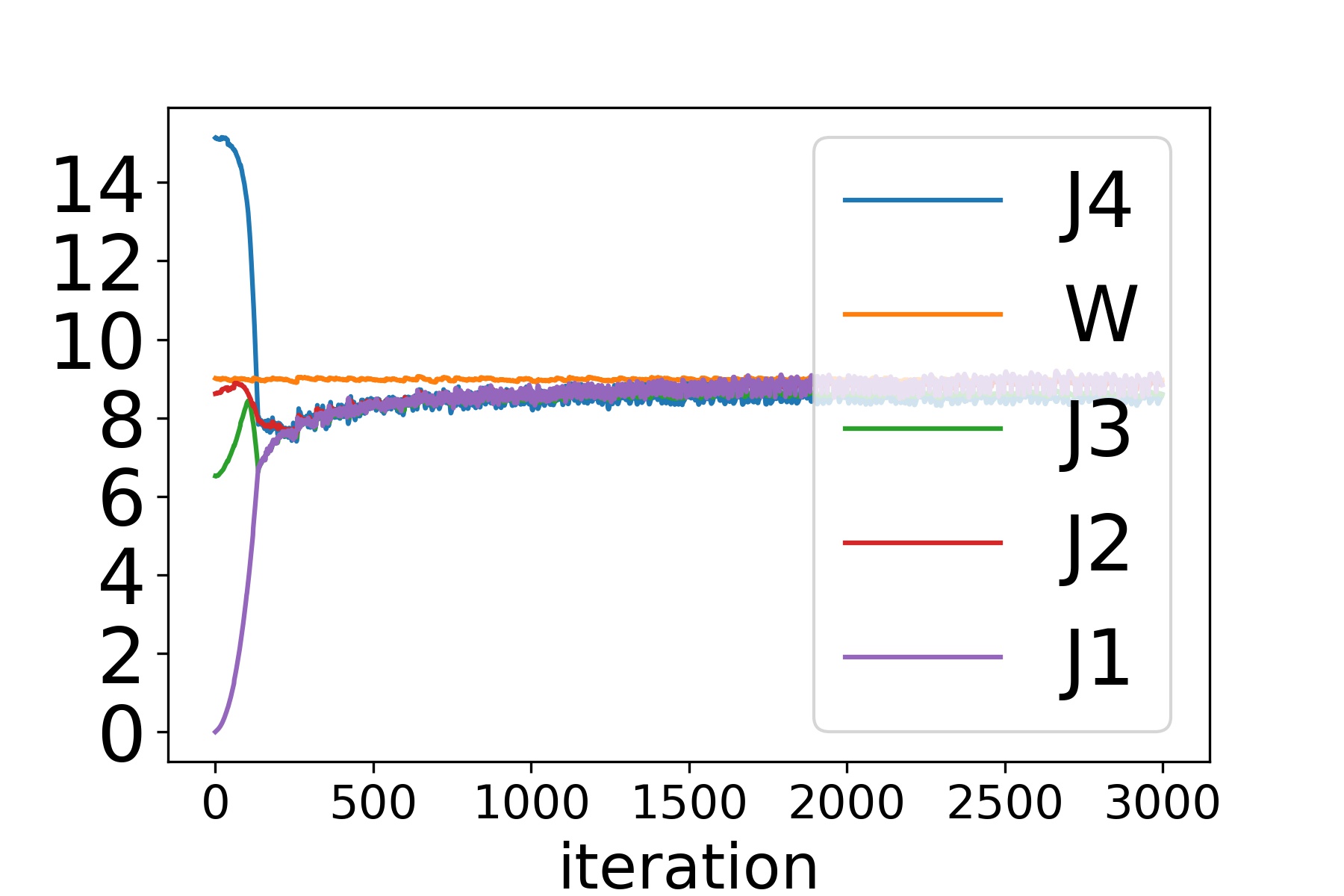}&
    \includegraphics[width=0.23\textwidth,clip=true,trim=0cm 0cm 0cm 1cm ]{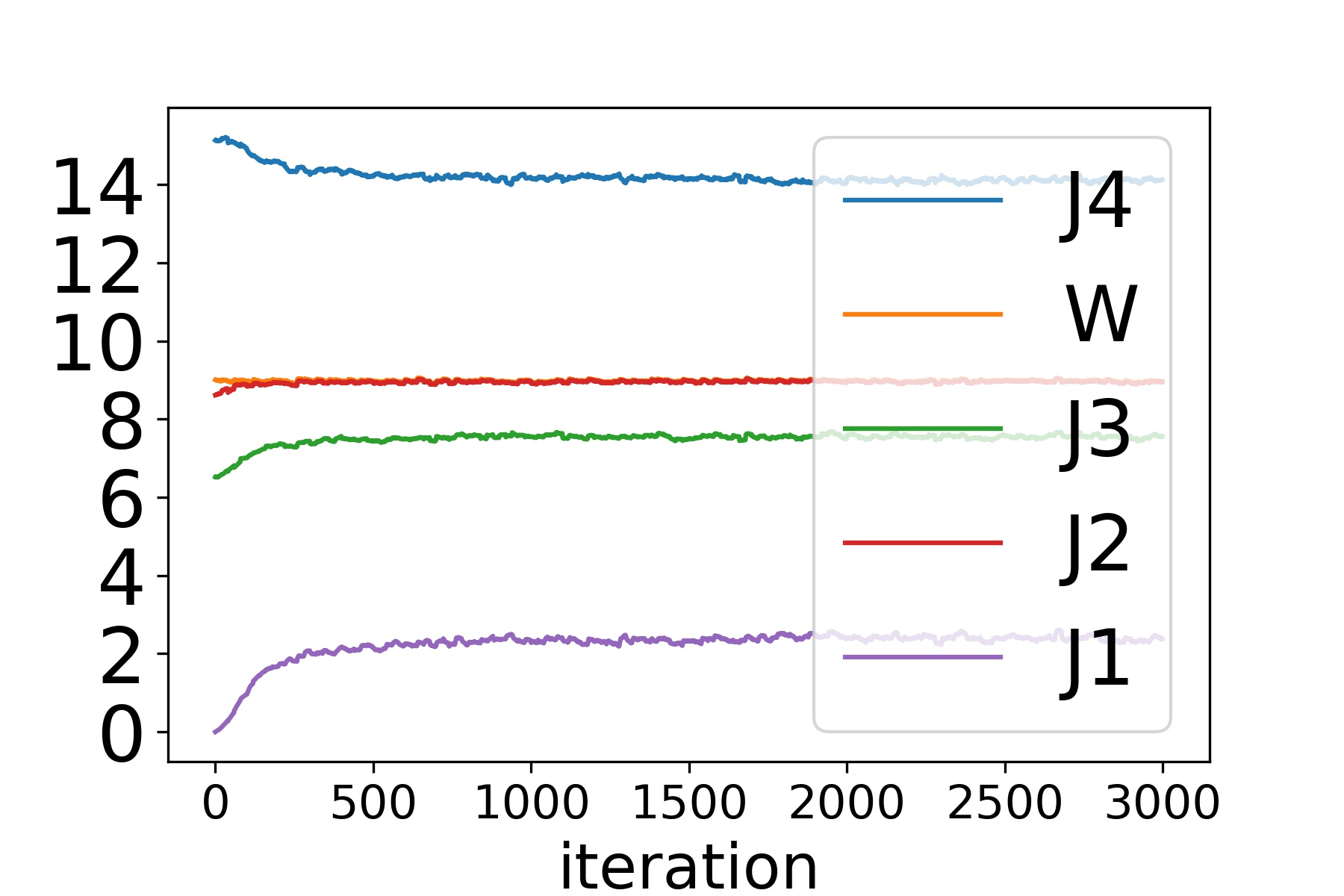}
    \\
    WGAN-GP, $\lambda=1$ & WGAN-GP, $\lambda=10$ \\ 
    \includegraphics[width=0.23\textwidth,clip=true,trim=0cm 0cm 0cm 1cm ]{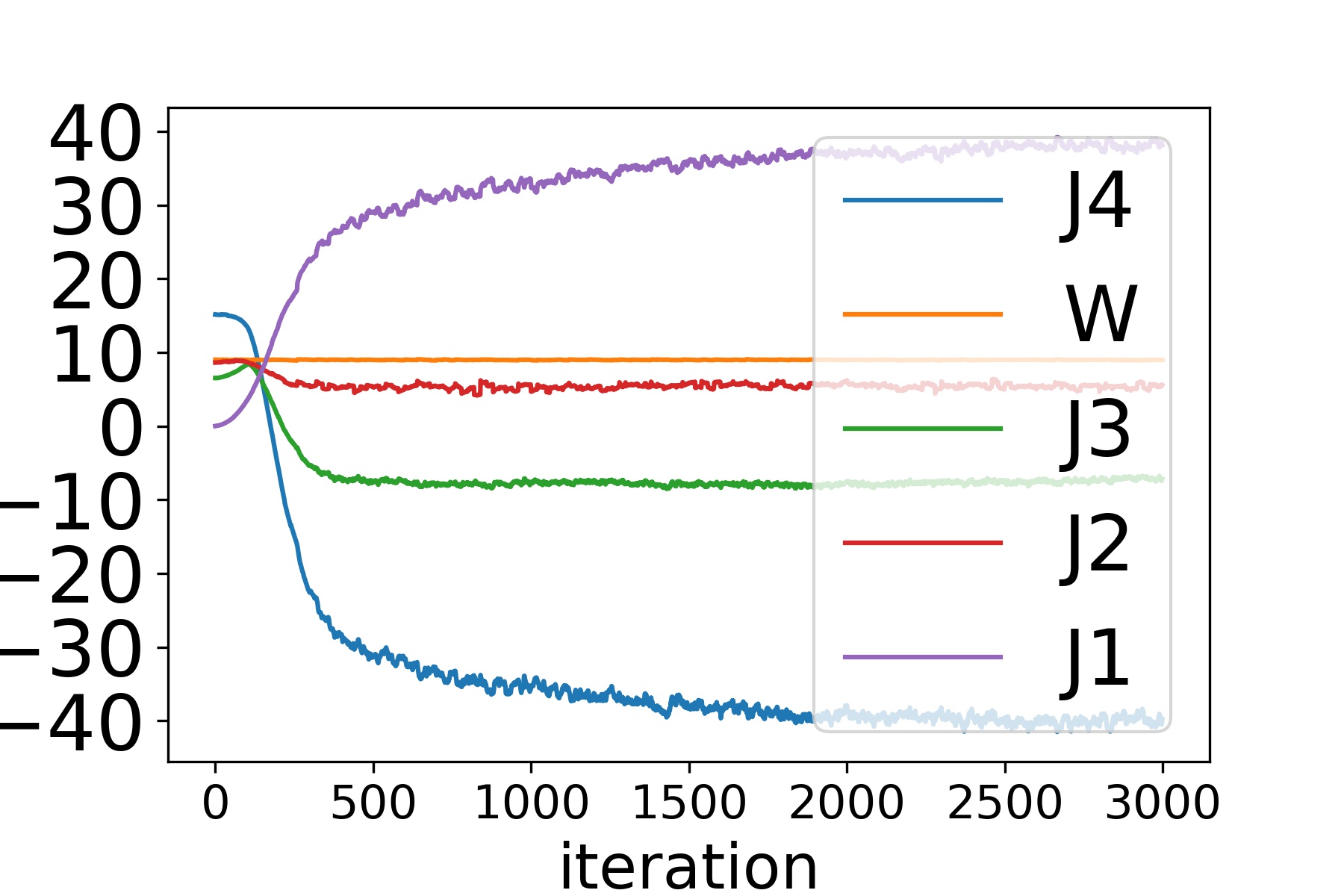}&
    \includegraphics[width=0.23\textwidth,clip=true,trim=0cm 0cm 0cm 1cm ]{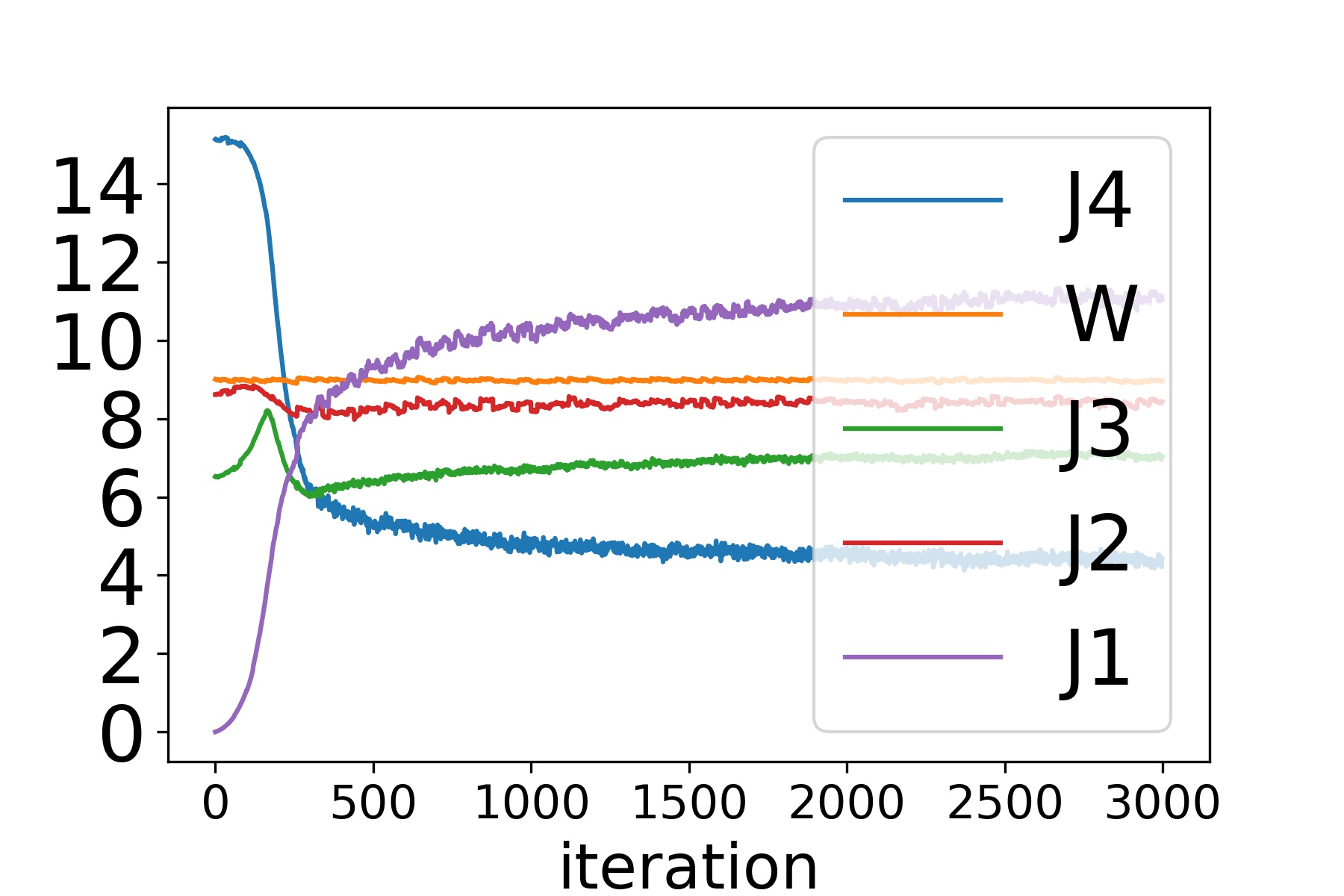}
    \end{tabular}
    \caption{The $\mathcal{J}_i$'s and the true Wasserstein distance (W) for the MNIST dataset.}
    \label{ot_mnist}
    \end{figure}

    In Figure~\ref{ot_mnist}, we sampled 5,000 images of digit 1 and 5,000 images of digit 2 from the MNIST dataset. 
    Applying Algorithm~\ref{al1}, we compute the Wasserstein distance between the two distributions $\mu$ and $\nu$ of 
    images of the digit 1 and digit 2, respectively. We observe that in our algorithm all objective functions $J_i$'s converge to the same value, the Wasserstein distance, while other algorithms do not feature such a convergence. Here, the Wasserstein distance ($W$) is computed using the POT Python Library \citep{flamary2021pot}. 
    
\section{Learning Generative Models}
In this section we apply our algorithm to generative learning in a WGAN framework. 

\subsection{Experimental Results}

	We present 
	results using the MNIST, Fashion MNIST and CIFAR-10 datasets. 
	We use the algorithm proposed in the previous section to generate synthetic images. 
	We find that the training procedure is more stable compared with 
	WGAN-GP in terms of the Lipschitz estimate while generating images of similar quality. In particular, 
	the proposed machinery is robust to hyperparameter tuning as the algorithm does not involve any parameters such as weight $\lambda$ or $n_{critic}$ which represents the number of discriminator update per each generator update. The details on the used neural network architecture 
	are provided in the supplementary material.
	
	We first examine the Lipschitz constant with real and synthetic data points. 
	To this end, we first sample 64 points from each real and synthetic datasets and compute the maximum slope of the discriminator within them, i.e. 
	$\text{max}_{{x} \sim \mu,{y} \sim \nu}|\frac{\phi(x)-\phi(y)}{|x-y|}|$. 
	The result is shown in Figure \ref{fig:lip}. It is interesting to observe that the Lipschitz constant is forced to be one quite well (left part of the figure).
	On the other hand, WGAN-GP underestimates both quantities. This is mainly because the gradient penalty term unnecessarily forces 
	the gradient of points to be one even if the norm of their gradients does not have to be equal to one. 

    Furthermore, our algorithm is successful when it comes to generating synthetic images without computing the gradients explicitly as can be seen in Figure \ref{fig:img}. 
    Our algorithm is remarkably faster than WGAN-GP while generating quality synthetic images. On average, our algorithm is six times faster than WGAN-GP as measured in wall-clock time. 
    We note that CoWGAN only needs to update the discriminator once per generator update while in WGAN-GP one usually needs to update the discriminator several times per generator update, which is given by an oracle. However, the main reason for the computational savings is that CoWGAN does not need to compute the gradients of the discriminator. Even if we set WGAN-GP to update discriminator once per each generator update, it is observed that our algorithm is faster than WGAN-GP. More specifically, it takes 4.2 seconds per 100 generator updates for CoWGAN while it takes 5 seconds for WGAN-GP on average with the same learning rate. 
    %

    Another observation is that infusing randomness in the updates of $\mathcal{J}_2$ can be beneficial. We update $\mathcal{J}_2$ with a fifty percent chance whenever it is less than $\mathcal{J}_1$. We call this modification 
    CoWGAN-P, where P indicates probability. 
    By introducing 
    $p$, we observed that 
    $\mathcal{J}_2$ 
    is updated more in tandem with $\mathcal{J}_3$, which 
    further improves 
    stability. Details of this experiment can be found in the supplemental materials.


\begin{figure}[t]
\centering

\begin{tabular}{cc}
Lipschitz constant & FID score \\  
\includegraphics[width=0.2\textwidth ]{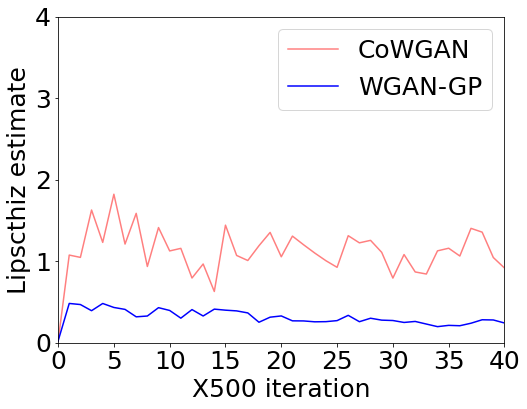}&
\includegraphics[width=0.2\textwidth ]{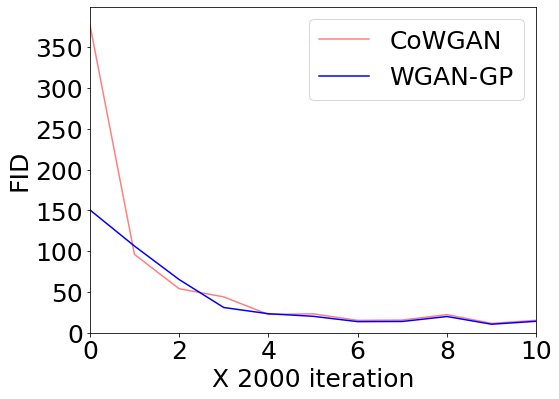}
\\
\includegraphics[width=0.2\textwidth ]{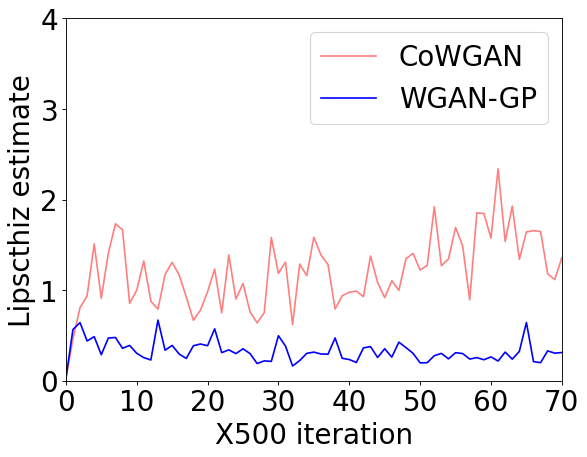}&
\includegraphics[width=0.2\textwidth ]{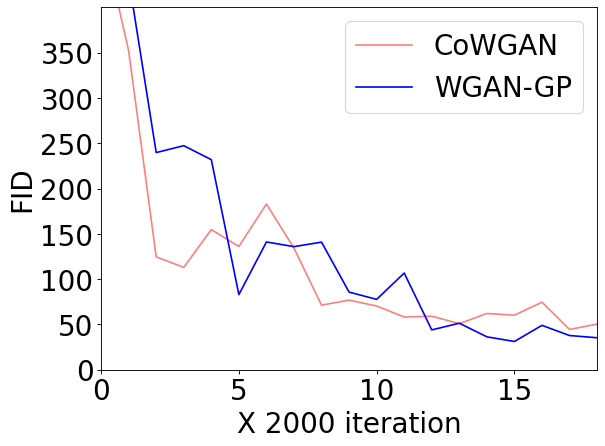}
\\
\includegraphics[width=0.2\textwidth ]{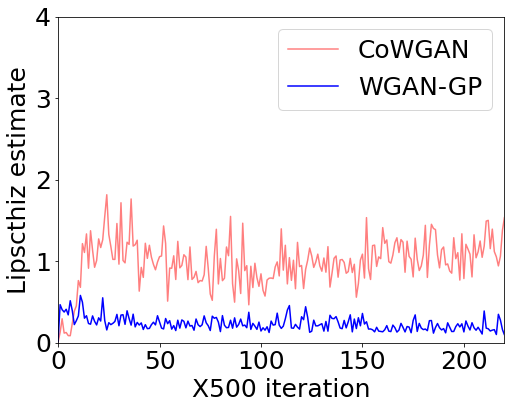}&
\includegraphics[width=0.2\textwidth ]{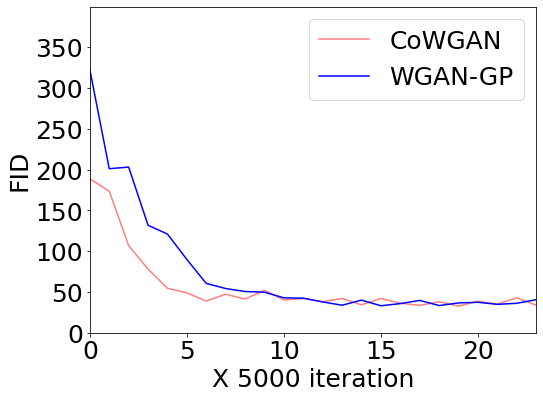}

\end{tabular}

\caption{MNIST (top), F-MNIST(middle) and CIFAR10 (bottom).}
\label{fig:lip}

\end{figure}







\begin{figure}[t]
\centering

\begin{tabular}{cc}

WGAN-GP & CoWGAN (ours) \\ 
\includegraphics[width=0.2\textwidth ]{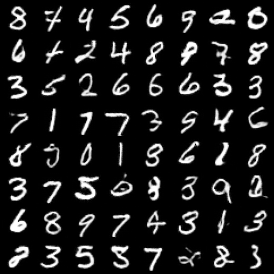}& 
\includegraphics[width=0.2\textwidth ]{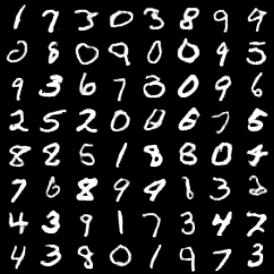}\\
\includegraphics[width=0.2\textwidth ]{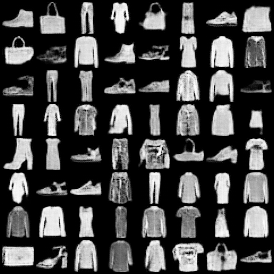}& \includegraphics[width=0.2\textwidth ]{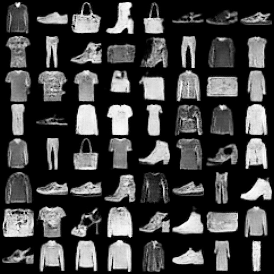}\\ 
\includegraphics[width=0.2\textwidth ]{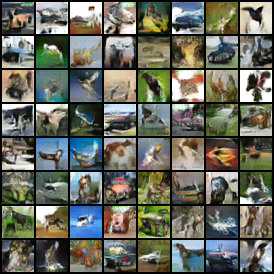}& \includegraphics[width=0.2\textwidth ]{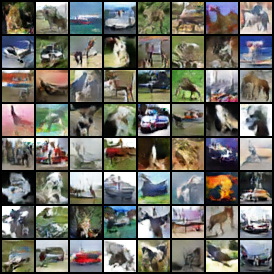}\\

\end{tabular}

\caption{Images on the left are generated by using WGAN-GP; images on the right are generated using our CoWGAN algorithm. From top to bottom the training data was MNIST, F-MNIST, and CIFAR-10. 
Visually, the generated images are of similar quality, but our algorithm runs six times faster in wall-clock time.}
\label{fig:img}

\end{figure}



\section{Conclusion}
In this work, we presented a novel algorithm that 
accurately estimates the Wasserstein distance between two probability measures 
and can be used to train generative models. 
The method exploits a combination of objective functions inspired in the recently proposed back-and-forth method and comes with a sound theoretical motivation based on optimal transport duality and a few new inequalities. 
Our method does not require explicit computation of the gradients of the discriminator in order to implement the Lipschitz constraint on the optimal discriminators in WGANs so that it is significantly faster than existing methods that rely on computation of these gradients. 
An added benefit of the method is that 
without gradient penalties, no corresponding hyperparameter tuning is needed. 

\section*{Acknowledgement}
The authors would like to thank Jinyoung Choi, Alpar Meszaros, Kangwook Lee and Sangwoong Yoon for helpful comments and insightful discussions.
Insoon Yang and Yeoneung Kim are supported by Brain Korea 21 and Institute of Information \& communications Technology Planning \& Evaluation (IITP) grant funded by the Korea government(MSIT) (No. 2020-0-00857, Development of cloud robot intelligence augmentation, sharing and framework technology to integrate and enhance the intelligence of multiple robots)


\bibliography{ref.bib}
\appendix

%

%

\onecolumn
\aistatstitle{Supplementary Materials}



\section{Proof of Proposition 1}






\begin{proposition}
\label{prop:1}
Assume that $d=n=1$ and $\mu \in \mathcal{P}_m(\Omega)$ for $m>1$. Then, for any median $y$ of $\mu$, $\nu = \delta_{y}$ is a global minimizer of 
\begin{align}
    \label{min3}
    \inf_{\nu \in P(\Omega)} \mathbb{E}_{\mu_n \sim \mu, \nu_n \sim \nu} [ W_1(\mu_n, \nu_n)].
\end{align} 
Here, a median of a probability measure $\mu$ is defined as a real number $k$ satisfying $\mu((-\infty, k]) \geq \frac{1}{2}$ and $\mu([k, \infty)) \geq \frac{1}{2}$.
\end{proposition}

\textit{Proof.} For simplicity, assume that $\mu = \frac{1}{m} \sum_{i=1}^m \delta_{X_i}$ and $X_1 < X_2 < \cdots < X_m$. Other cases can be handled similarly. We claim that if $y \in [X_{[(m+1)/2]}, X_{[(m+2)/2]}]$, then $\nu = \delta_{y}$ is a global minimizer of \eqref{min3}. 
As $d=n=1$, the objective function in \eqref{min3} can be represented as
$$ I := E_{Y \sim \nu} \left[ \sum_{i=1}^m |X_i - Y| \right].$$
From the triangular inequality,

\begin{align*}
I &=  \frac{1}{2} E_{Y \sim \nu} \left[ \sum_{i=1}^m |X_i - Y| + |X_{m+1-i} - Y| \right], \\
&\geq \frac{1}{2} \sum_{i=1}^m |X_i - X_{m+1-i}|.    
\end{align*}

On the other hand, if $\nu = \delta_{y}$ for $y \in [X_{[(m+1)/2]}, X_{[(m+2)/2]}]$, then the objective function $E_{Y \sim \nu} \left[ \sum_{i=1}^m |X_i - Y| \right]$ equals to 
\begin{align*}
&\sum_{i=1}^{[(m+1)/2]} (y - X_i) +  \sum_{i=[(m+2)/2]}^{m} (X_i - y)\\ 
&= \sum_{i=1}^{[(m+1)/2]} (X_{m+1-i} - X_i).
\end{align*}
As the minimum of \eqref{min3} attains at $\nu = \delta_{y}$, we conclude. \hfill$\Box$

$\,$

$\,$

$\,$

$\,$

$\,$

\section{Remarks on the objective functions}

	This Kantorovich formulation motivates us to consider the following four objective functions,
	\begin{align*}
		\mathcal{I}_1(\phi) &= \int_\Omega \phi \text{d}\mu + \int_\Omega (-\phi) \text{d}\nu, \\
		\mathcal{I}_2(\phi) &= \int_\Omega \phi \text{d}\mu + \int_\Omega \phi^c \text{d}\nu,\\
		\mathcal{I}_3(\phi) &= \int_\Omega (-\phi)^c \text{d}\mu + \int_\Omega (-\phi) \text{d}\nu,\\
		\mathcal{I}_4(\phi) &= \int_\Omega (-\phi)^c \text{d}\mu + \int_\Omega \phi^c \text{d}\nu.
	\end{align*}
	We point out that if $\phi$ is a 1-Lipschitz function, then $\phi^c = -\phi$ and thus
	\begin{align}
		\label{eqn:ieq}
		\mathcal{I}(\phi) = \mathcal{I}(\phi) = \mathcal{I}(\phi) = \mathcal{I}(\phi) \hbox{ for any } \phi \in Lip_1(\Omega).
	\end{align}

As a consequence, we have the following relation between $\mathcal{I}_i$s with appropriate admissible sets and the Wasserstein distance,
	\begin{align*}
		\begin{split}
		\sup_{\|\phi\mathcal{I}_{Lip} \leq 1} \mathcal{I}_1(\phi) = \sup_{\phi \in C(\Omega)} \mathcal{I}_2(\phi) = \sup_{\phi \in C(\Omega)} \mathcal{I}_3(\phi) = \sup_{\phi \in C(\Omega)} \mathcal{I}_4(\phi) = W_1(\mu,\nu).
	    \end{split}
	\end{align*}

\section{Neural network Architectures and learning rates}
For CoWGAN, we use Adam optimizer with $\beta_1=0.5$, $\beta_2=0.999$. For the learning rate, $5\cdot 10^{-5}$ and $10^{-4}$ are used for the discriminator and the generator. In contrast, for the learning rate, $10^{-4}$ is used for both discriminator and generator in WGAN-GP. 
\begin{table}[h]
\parbox{.3\linewidth}{
\begin{tabular}{ |c| } 
 \hline
 Discriminator \\ 
 \hline
Input: 1*32*32\\
4*4 conv. 256, Pad = 1, Stride = 2, lReLU 0.2\\
4*4 conv. 512, Pad = 1, Stride = 2, lReLU 0.2\\
4*4 conv. 1024, Pad = 1, Stride = 2, lReLU 0.2\\
4*4 conv. 1, Pad = 0, Stride = 1\\
Reshape 1024*4*4 (D)\\
 \hline
 Generator \\
 \hline
Input: Noise z 100 \\
4*4 deconv. 1024, Pad = 0, Stride = 1, ReLU \\
4*4 deconv. 512, Pad = 1, Stride = 2, ReLU \\
4*4 deconv. 256, Pad = 1, Stride = 2, ReLU \\
4*4 deconv. 1, Pad = 1, Stride = 2, ReLU \\
\hline
\end{tabular}
\caption {MNIST~and~F-MNIST}
\label{table:1}
}
\ \ \ \ \ \ \ \ \ \ \ \ \ \ \ \ \ \ \ \ \ \ \ \
\parbox{.3\linewidth}{
\begin{tabular}{ |c| } 
 \hline
 Discriminator \\ 
 \hline
Input: 1*32*32\\
4*4 conv. 256, Pad = 1, Stride = 2, lReLU 0.2\\
4*4 conv. 512, Pad = 1, Stride = 2, lReLU 0.2\\
4*4 conv. 1024, Pad = 1, Stride = 2, lReLU 0.2\\
4*4 conv. 1, Pad = 0, Stride = 1\\
Reshape 1024*4*4 (D)\\
 \hline
 Generator \\
 \hline
Input: Noise z 100 \\
4*4 deconv. 1024, Pad = 0, Stride = 1, BatchNorm2d, ReLU \\
4*4 deconv. 512, Pad = 1, Stride = 2, BatchNorm2d, ReLU \\
4*4 deconv. 256, Pad = 1, Stride = 2, BatchNorm2d, ReLU \\
4*4 deconv. 1, Pad = 1, Stride = 2, BatchNorm2d, ReLU \\
\hline
\end{tabular}
\label{table:1}
\caption{CIFAR10 }
}
\end{table}

\newpage
\section{CoWGAN and CoWGAN-P}
The comparison between CoWGAN and CoWGAN-P is provided. We see mixing the update steps improve the results. The algorithm for CoWGAN-P is given as

	\begin{algorithm}[H]
	\label{al0-p}
	\For{$iter$ of training iterations}{
		\For{$t = 1, 2, \dots, N_{critic}$, $p\in$ Unif(0,1)}{
			\uIf{$\mathcal{J}_2 < \mathcal{J}_1 \text{ and } p<0.5$}{
				$\eta \leftarrow \hbox{Adam}(-\mathcal{J}_2, \eta)$
			}
			\uElseIf{$\mathcal{J}_3 < \mathcal{J}_1$}{
				$\eta \leftarrow \hbox{Adam}(-\mathcal{J}_3, \eta)$
			}
			\Else{
				$\eta \leftarrow \hbox{Adam}(-\mathcal{J}_1, \eta)$
				}
		}
	$\theta \leftarrow  \hbox{Adam}(\mathcal{J}_1, \theta)$  
	}	
		\caption{CoWGAN with probabilistic mixing}
		\label{alg:p}
	\end{algorithm}

\begin{figure}[h]
\centering

\begin{tabular}{cc}
\includegraphics[width=0.4\textwidth ]{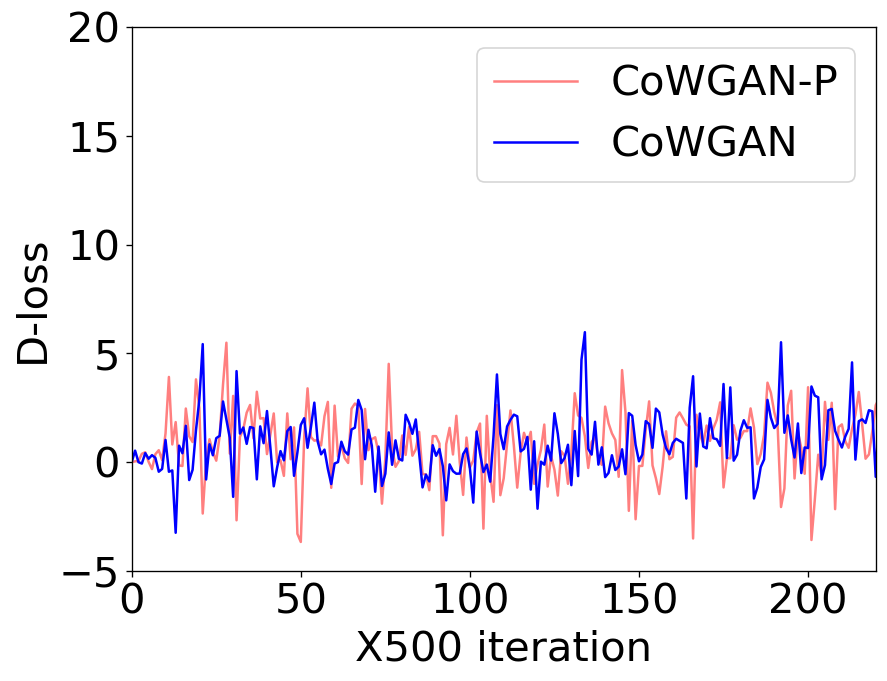} 
\includegraphics[width=0.4\textwidth ]{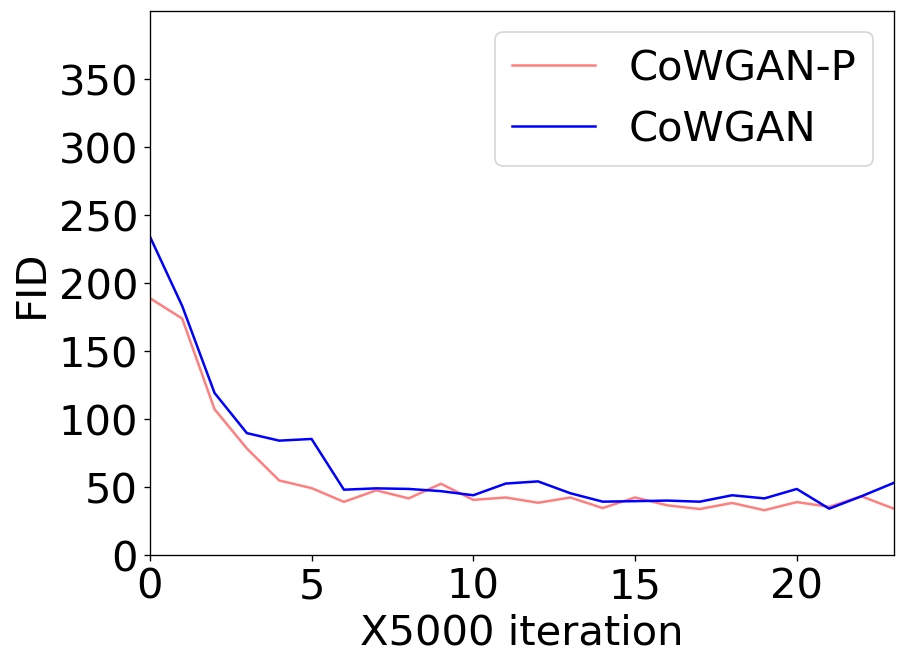} 
\end{tabular}

\caption{Lipschitz estimate within the real data (left), between real and synthetic data (right) for MNIST (top), F-MNIST (middle) and CIFAR10 (bottom)}
\label{fig:lip}

\end{figure}

\newpage
\section{Additional experimental result}






\subsection{Lipschitz estimate}

Here we provide additional experimental data, Lipschitz estimate within real data
$$\text{max}_{x,y \sim \mu} \left|\frac{\phi(x)-\phi(y)}{|x-y|}\right|$$ 
and Lipschitz estimate between real and synthetic data 
$$\text{max}_{{x} \sim \mu,{y} \sim \nu}\left|\frac{\phi(x)-\phi(y)}{|x-y|}\right|$$
together with learning curves for all training data used.

\begin{figure}[h]
\centering
\begin{tabular}{cc}
\includegraphics[width=0.3\textwidth ]{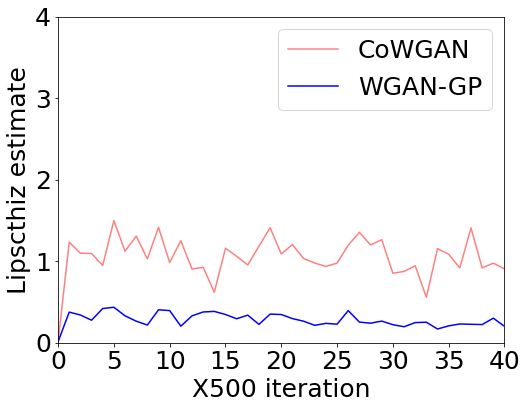} \includegraphics[width=0.3\textwidth ]{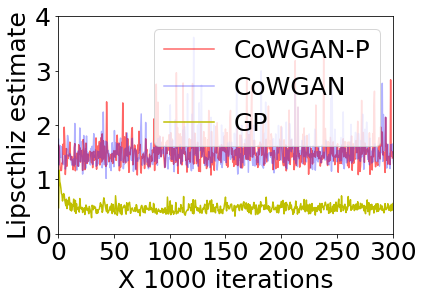}\\
\includegraphics[width=0.3\textwidth ]{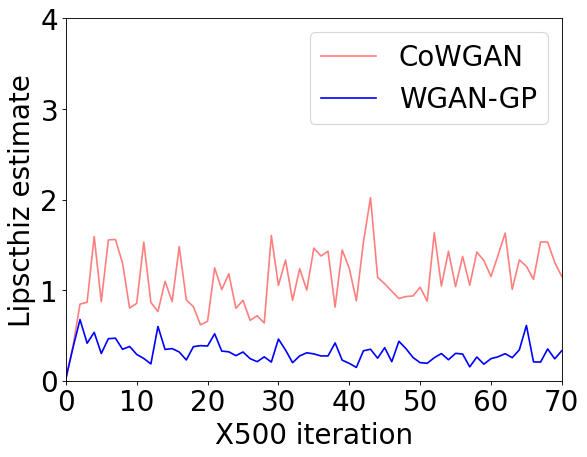} \includegraphics[width=0.3\textwidth ]{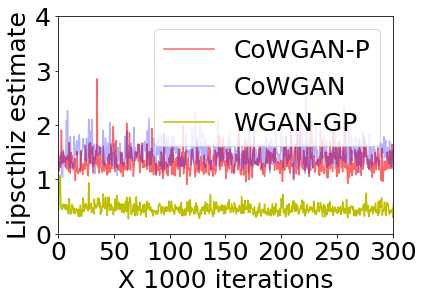}\\
\includegraphics[width=0.3\textwidth ]{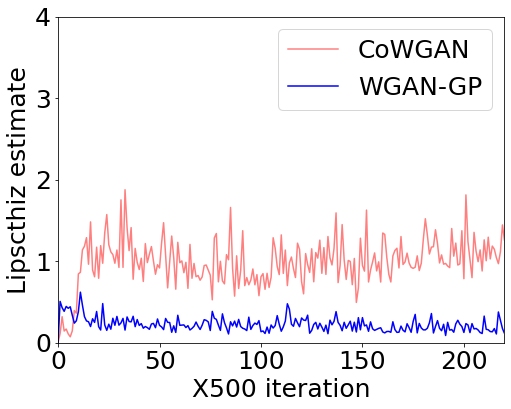} \includegraphics[width=0.3\textwidth ]{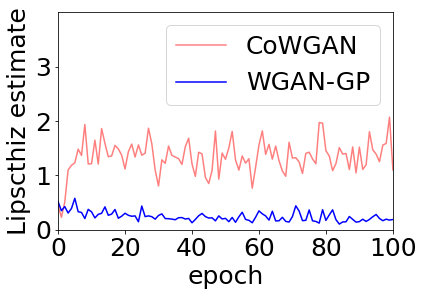} 

\end{tabular}

\caption{Lipschitz estimate within the real data (left), between real and synthetic data (right) for MNIST (top), F-MNIST (middle) and CIFAR10 (bottom)}
\label{fig:lip}

\end{figure}

\newpage

\subsection{Discriminator loss and FID score}

\begin{figure}[h]
\centering

\begin{tabular}{cc}

\includegraphics[width=0.4\textwidth ]{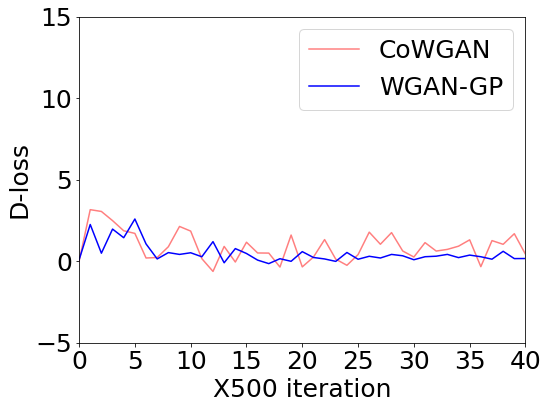}& \includegraphics[width=0.42\textwidth ]{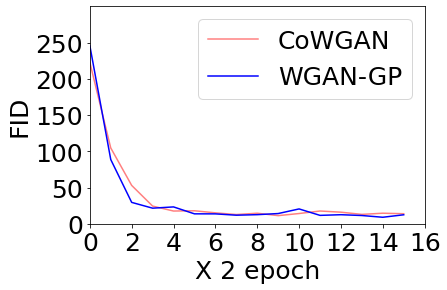}\\ 

\includegraphics[width=0.4\textwidth ]{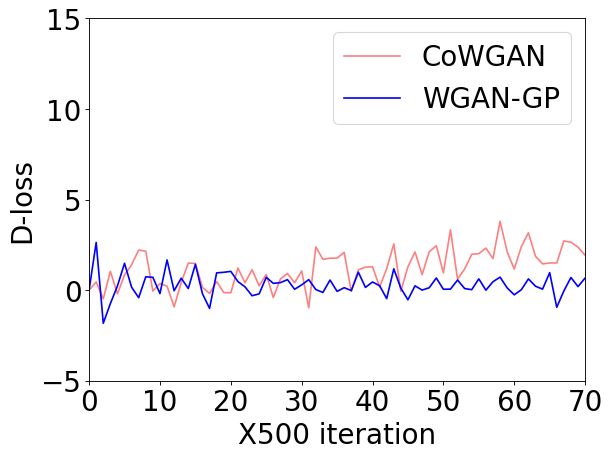}& \includegraphics[width=0.42\textwidth ]{F_FID.png}\\ 

\includegraphics[width=0.4\textwidth ]{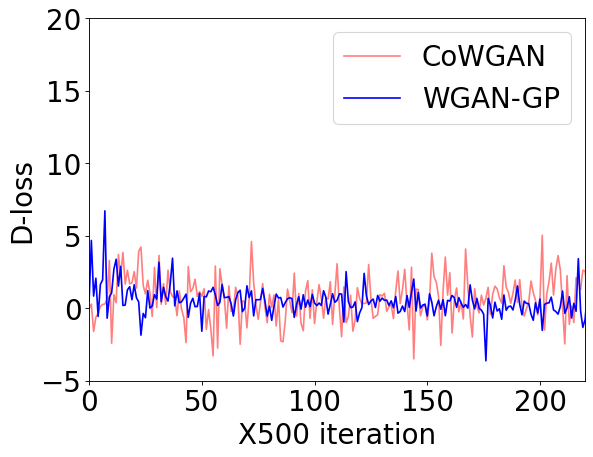}& \includegraphics[width=0.42\textwidth ]{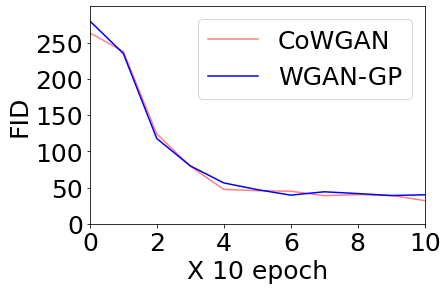}\\ 

\end{tabular}

\caption{Discriminator loss ($\mathcal{J}_1$) (left) and FID score (right) for MNIST (top), F-MNIST (middle) and CIFAR10 (bottom)}
\label{fig:curve}

\end{figure}

\end{document}